\documentclass[oneside,11pt]{article}
\usepackage{caption}
\usepackage{blindtext}

\usepackage[abbrvbib, preprint]{jcustom}
\usepackage{lastpage}
\usepackage{fullpage}

\usepackage[utf8]{inputenc} 
\usepackage[T1]{fontenc}           
\usepackage{booktabs} 
\usepackage{hyperref}

\hypersetup{
    colorlinks=true,
    allcolors=blue
}

\usepackage{amsfonts,amsmath,amsthm,amssymb}       
\usepackage{nicefrac}       
\usepackage{microtype}      
\usepackage{xcolor}         
\usepackage{pifont}
\usepackage{tikz}
\usepackage{subcaption}

\usetikzlibrary{arrows.meta}
\usepackage{wrapfig}
\usepackage[presets={vec-cev}]{letterswitharrows}
\usepackage{cleveref}
\AddToHook{cmd/appendix/before}{
    \crefalias{section}{appendix}
    \crefalias{subsection}{appendix}
}
\usepackage{graphicx}
\usepackage{amsfonts}
\usepackage{mathtools}
\usepackage{booktabs}
\usepackage{nicefrac}

\theoremstyle{remark}
\newtheorem{remark}{Remark}

\definecolor{src}{RGB}{76,114,176}
\definecolor{tgt}{RGB}{221,132,82}

\tikzset{
  traj/.style={black!75, line width=0.9pt, -{Latex[length=1.8mm]}},
  noisytraj/.style={black!70, line width=0.8pt, -{Latex[length=1.6mm]}, opacity=0.75},
  margsrc/.style={fill=src!12, draw=src!55, line width=0.5pt},
  margtgt/.style={fill=tgt!12, draw=tgt!55, line width=0.5pt},
  sp/.style={fill=src, draw=src},
  tp/.style={fill=tgt, draw=tgt},
}

\title{Model reduction in time: Learning flow maps for reducing stochastic dynamical systems in time}

\author{%
 \name Shreya Jha \\
  \addr Courant Institute School of Mathematics, Computing, and Data Science, New York University\\
  \AND
  Benjamin Peherstorfer\\
   \addr Institute of Mathematics, EPFL
}

\begin{document}

\maketitle

\begin{abstract}
Predictions with dynamical systems can be computationally expensive---even with reduced models---because reaching future prediction times can require many sequential time integration steps. We propose Flow-Map First-Order Trajectory Matching (FM-FTM), a nonintrusive model reduction approach that accelerates ensemble predictions of stochastic dynamical systems by learning flow maps that replace multiple time-integration steps. Our method uses first-order trajectory matching (FTM) to learn a deterministic representation of the stochastic probability-current dynamics and then reduces its evolution over time into a flow map whose evaluation directly yields predictions over long time horizons. Numerical experiments on evolution problems with stochastic forcing demonstrate that FM-FTM accelerates ensemble predictions by up to two orders of magnitude while accurately predicting time marginals and ensemble statistics.
\end{abstract}

\section{Introduction}
Ensemble predictions of stochastic and chaotic dynamical systems are essential for uncertainty quantification and probabilistic forecasting.
Accurate numerical simulations can generate the trajectories needed for these predictions but the computational costs are often too high when the number of ensemble members is large. 
Reduced models can decrease the computational costs by restricting the state space of underlying dynamical systems to a lower dimensional subspace~\cite{rozza2008reducedbasis,benner2015survey,antoulas2020interpolatory,benner2021morvolume1,benner2021morvolume2,kramer2024learning,kramer2026survey,hesthaven2026nonlinear}.
However, reducing the dimension of the states reduces the cost per time step but not necessarily the number of time steps. Even when each time step with a reduced dynamical-system model is inexpensive, the sheer number of time steps can still incur high costs. 

In this work, we reduce the cost of time integration itself
by learning a flow map whose evaluation replaces multiple
sequential integration steps.
We build on first-order trajectory matching (FTM)~\cite{jha2026ftm}, which learns a deterministic velocity field representing the probability-current dynamics of a stochastic process.
Rather than sequentially integrating the corresponding dynamical-system model in the online phase, we approximate its evolution via a
flow map that directly predicts the state over far time horizons.
The learned flow map enables far-horizon predictions without time integration and thus avoids having to make many time integration steps.

Reducing the computational cost of time integration
has a long history in numerical analysis and model reduction. First, there are time-parallel methods such as parareal \cite{lions2001parareal} and multigrid reduction in time \cite{falgout2014mgrit}, which reduce the wall-clock time of time integration. 
Related parallel-in-time formulations have also been
explored for chaotic turbulent flows
~\cite{wang2013parallel}.
Other approaches exploit fine- and coarse-scale descriptions to accelerate time integration, including for stochastic ensembles; see, e.g.,~\cite{gear2003projective,kevrekidis2003equationfree,DebrabantSamaeyZielinski2017MicroMacro}. Data-driven coarse-graining approaches learn effective
evolution equations from simulation data, including equations
for probability distributions of stochastic systems
~\cite{bakarji2021datadriven}.

There also is a body of literature on reducing the costs of time integration of reduced models. The work 
~\cite{lemaitre2010equationfree} proposes an equation-free
reduced model that approximates the evolution of
proper-orthogonal-decomposition coefficients over a coarse
time step using an explicit polynomial map.
Temporal forecasting~\cite{carlberg2015temporalforecast}
reduces the cost of solving time-discrete reduced models,
while data-driven coarse propagators \cite{carlberg2019timeparallel} accelerate time-parallel
integration. Space-time projection methods
~\cite{choi2019spacetime,choi2021spacetime} reduce the time-integration costs 
by approximating the solution in a low-dimensional
space-time subspace, rather than reducing only the spatial
dimension. Certified and windowed space-time reduced models
offer further accelerate for reducing temporal
complexity
\cite{Yano2014SpaceTime,ParishCarlberg2021Windowed,
ShimizuParish2021Windowed}, while some nonintrusive
deep-learning reduced models can directly evaluate solutions
at selected times
\cite{fresca2021comprehensive,FrescaManzoni2022PODDLROM}. Dynamic mode decomposition provides another data-driven
approach to temporal prediction, with error estimates
for extrapolation in time~\cite{lu2020prediction}
and extensions to inhomogeneous dynamical systems
~\cite{lu2021extended}. Closer to our approach is work on flow-map learning, which approximates discrete-time evolution operators directly from trajectory data, for example using residual neural networks \cite{qin2019datadriven}; see also \cite{churchill2023flowmap} for an overview of flow-map learning in this context. The work \cite{chen2025due} provides
a broader framework and software tools for
data-driven learning of unknown dynamical systems using ideas related to flow maps. Flow maps also have been learned over variable time intervals and with encouraging consistency with the
semigroup property during training~\cite{chen2023deeposg}. Also with the objective to acclerate time integration, the work \cite{liu2022hierarchical} proposes hierarchical time-steppers that learn flow maps over different physical-time intervals and combine them to accelerate long-horizon predictions. The approach \cite{liu2022hierarchical} reduces the number of sequential time-stepping evaluations by exploiting learned evolution maps at multiple temporal resolutions. 
However, these deterministic flow-map approaches
do not directly address the additional challenge
of preserving probability distributions and
statistical quantities of interest (QoIs) when predicting
stochastic ensembles.

Our approach specifically builds on ideas developed to accelerate sampling in diffusion and flow-based
generative models~\cite{song2021score,lipman2023flow}.
These models generate samples by numerically integrating
learned stochastic or deterministic dynamics, which can
require many sequential evaluations. Progressive
distillation~\cite{salimans2022progressive} reduces the
number of sampling steps by distilling multiple teacher
steps into fewer student steps. Consistency models
~\cite{song2023consistency} learn mappings along
generative trajectories that enable sampling with one
or a few model evaluations.
A building block for our work is flow-map matching
~\cite{boffi2025flowmapmatching} and related MeanFlow and flow-map self-distillation
~\cite{geng2025meanflow,boffi2025selfdistillation}, which learns the 
flow maps associated with generative velocity fields,
including through distillation of a learned velocity
model. 
Together, these methods demonstrate that 
flow maps can replace numerical integration of learned
velocity fields with substantially fewer evaluations.
While the generative approaches above reduce integration in artificial sampling time, our focus is on evolution in the physical time of a dynamical system. 

\begin{figure}[t]
    \centering
    \includegraphics[width=1.0\columnwidth]{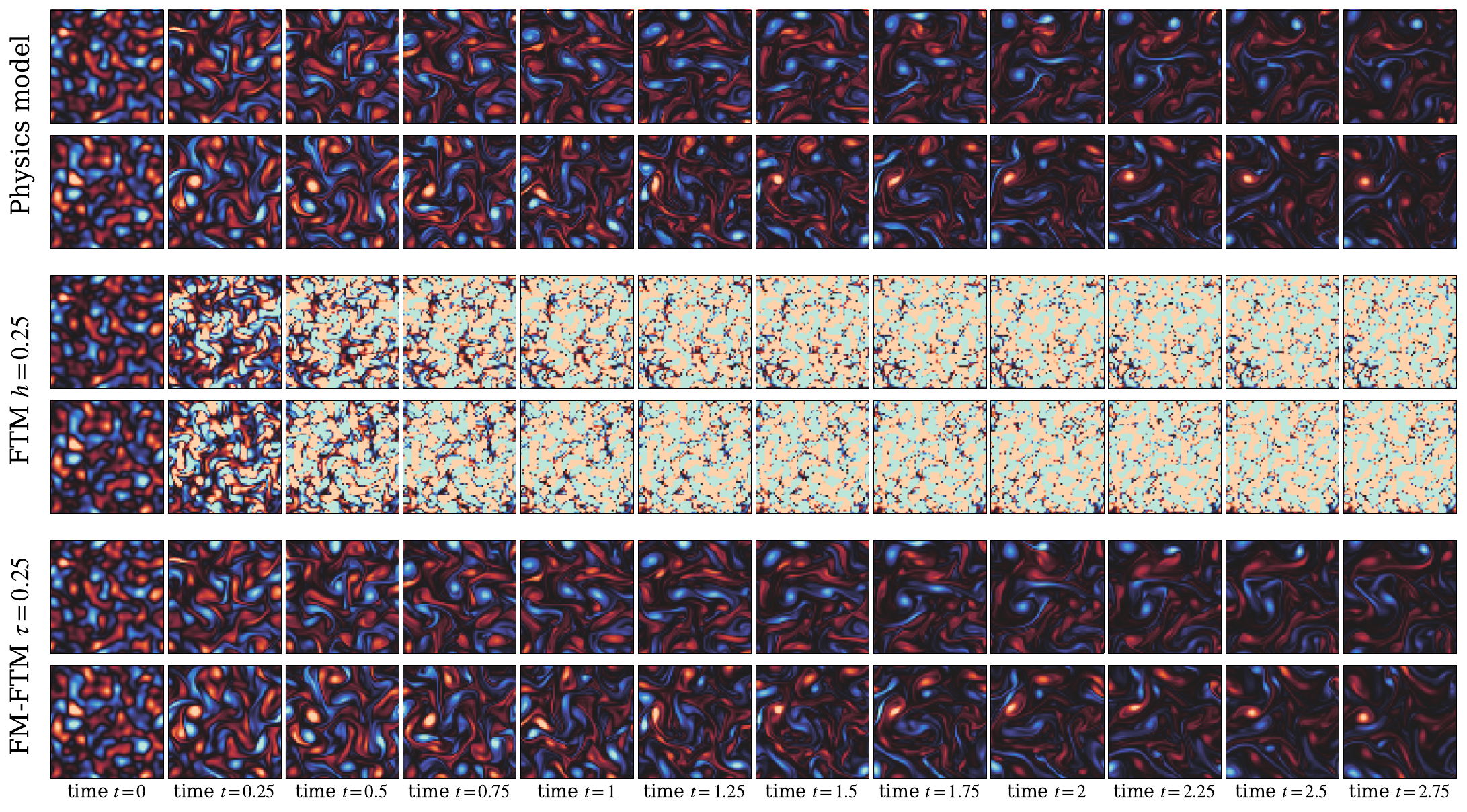}
    \caption{While FTM quickly loses accuracy at coarse time steps, FM-FTM maintains accurate predictions by taking far-horizon time jumps, replacing many sequential integration steps with a single flow-map evaluation and thereby accelerating ensemble prediction.}
    \label{fig:navier_stokes_grid}
\end{figure}

Flow maps have also been learned for stochastic systems in physical time. The authors of \cite{chen2024stochastic} combine deterministic and generative stochastic components to learn fixed-step stochastic transitions, while
the work \cite{LiuChenXiuZhang2025ConditionalDiffusion} learns such transitions using supervised data generated by a reverse probability-flow ordinary differential equation (ODE). Both approaches generate long-horizon predictions through repeated
applications of a learned stochastic map, in contrast to our work that requires a single neural-network evaluation per long-horizon prediction. There are data-driven approach that capture time-marginal dynamics and require only a single neural-network evaluation per time step \cite{neklyudov2023actionmatching,berman2024hoam,blickhan2025dice,schwerdtner2026twoparameter}, in contrast to traditional auto-regressive diffusion- and flow-based models \cite{chen2024probforecast,cachay2023dyffusion,KOHL2026108641}. Other data-driven approaches have  enabled large physical-time steps in stochastic simulation by inferring numerical integrators adapted to coarse time steps~\cite{LiLuYe2022ISALT,LiLuTaoYe2023NySALT} or learning conditional stochastic collocation points for Monte Carlo sampling~\cite{LiuGrzelakOosterlee2022SevenLeague}. 
There is also work on conditional normalizing flows, which have been used to directly sample stochastic states at prescribed physical times without intermediate time integration \cite{YangEtAl2024Pseudoreversible}. In contrast, our approach learns a single variable-horizon flow map that transports stochastic ensembles between arbitrary pairs of physical times without modeling the full stochastic transition law.

We approach the learning of flow maps for ensemble prediction in two stages. First, we use FTM to learn a deterministic dynamical-system model that mimics the stochastic dynamics in the sense that it preserves the time marginals and local probability mass transport. Second, we approximate
the evolution of the resulting deterministic dynamical system with a flow map, which leads to our approach called Flow-Map FTM (FM-FTM).
This separates the representation of stochastic dynamics from the reduction of their computational cost in time. Unlike conventional coarse time-steppers with a fixed step size, our FM-FTM can predict the ensemble evolution over different time horizons. Thus, the temporal resolution in the online phase  is not restricted to a prescribed coarse time step. 
Furthermore, FM-FTM is nonintrusive in the sense that it requires only 
trajectory data and no access to the governing equations or corresponding discretized operators.  In this respect, it follows the broader and long line of nonintrusive model reduction works; see, e.g., \cite{peherstorfer2016data,guo2018reduced,regazzoni2019machine,qian2020lift,lu2021deeponet,li2021fno,fresca2021comprehensive,guo2022bayesian,regazzoni2024latent,kramer2024learning,diaz2026kernelrom}. Our numerical results show that the learned flow maps accelerate ensemble predictions by at least one order of magnitude while accurately predicting ensemble statistics and QoIs that depend on the probability current of the stochastic process.

\section{Preliminaries and problem formulation}
We provide preliminaries and state the problem formulation. 
\subsection{Trajectory data and stochastic dynamics}
\label{sec:trajectory_data}
Let $(X(t))_{t\in[0,T]}$ be a continuous-time stochastic
    process with state space $\mathbb{R}^d$.
    We are given $N$ trajectories observed at $K+1$
    equally spaced times,
    \[
    0 = t_0 < t_1 < \dots < t_K  = T\,.
    \]
    Writing $X^{(i)}$ for the $i$th trajectory, the
    training dataset is
    \begin{equation}
    \label{eq:TrainingData}
    \mathcal{X}
    =
    \left\{
    X^{(i)}(t_k)
    \,\middle|\,
    i=1,\ldots,N,\;
    k=0,\ldots,K
    \right\}.
    \end{equation}
    We assume that $X(t)$ has a probability density
    $\rho(t):\mathbb{R}^d\to[0,\infty)$.
    We refer to the distribution $\rho(t)$ of the state at a single time $t$
    as the time marginal.
    Furthermore, $\rho(0)$ is the density of the
    initial condition.

    We consider stochastic processes governed by stochastic differential equations of
    the form
    \begin{equation}
    \label{eq:stochastic_dynamics}
    \mathrm{d}X(t)
    =
    b(t,X(t))\,\mathrm{d}t
    +
    A(t)\,\mathrm{d}W(t),
    \qquad
    X(0)\sim\rho(0)\,,
    \end{equation}
    where $b:[0,T]\times\mathbb{R}^d\to\mathbb{R}^d$
    is a smooth drift, $A:[0,T]\to\mathbb{R}^{d\times d}$
    is a smooth diffusion coefficient, and $W$ is a
    standard $d$-dimensional Wiener process.
    The drift specifies the deterministic contribution
    to the dynamics, while $A$ determines how the random
    forcing enters the state.
    Our learning procedure uses the trajectory data
    \eqref{eq:TrainingData} and does not require $b$ or $A$.

\subsection{Problem formulation}
\label{sec:problem_formulation}
We seek to rapidly generate ensembles to estimate QoIs as well as statistics of the distributions $\rho(t)$ at selected prediction times. 
Generating an ensemble by time stepping a dynamical model with a fixed integration time step $h>0$ up to time $T$ requires at least $T/h$ sequential steps, which typically incurs at least $T/h$ many function evaluations. 
 This applies both to the numerical simulation of \eqref{eq:stochastic_dynamics}, when its coefficients
    are available, and to dynamical models learned from data.
    These intermediate steps are required by the
    integrator even when predictions are requested only at times separated by intervals of length $\tau \gg h$ as shown in \Cref{fig:navier_stokes_grid}. 
A reduced dynamical model or  learned surrogate can lower the cost of each time step but does typically not reduce the number of steps. 

Our objective is to advance ensembles over horizons $\tau\gg h$ without carrying out all the intermediate time integration steps, while retaining accurate statistical estimates of interest.
    We thus target reducing the number of sequential time steps required for ensemble prediction, rather than only the cost per time step.

\section{Flow maps for first-order trajectory matching}
\label{sec:flow_maps}
We aim to learn the effect of time integration once and reuse it for subsequent ensemble predictions to jump over large time horizons $\tau \gg h$, instead of performing many small time steps $h$.
For the stochastic systems considered here, our construction first obtains a deterministic model of the ensemble evolution. 
    We build on the probability-current velocity, which describes
the local transport of probability mass through state space
~\cite{nelson1967dynamical,chetrite2009eulerian,
pavliotis2014stochastic}. In particular, we learn the probability-current velocity with FTM from the training data \eqref{eq:TrainingData} as in \cite{jha2026ftm}.
We then learn the corresponding flow map, which  assigns a state $x$ at time $t$ to the
    state reached at time $t + \tau$ under the deterministic
    model. Once learned, one flow map evaluation can jump over time horizons $\tau \gg h$, without integrating through the intermediate times.

\subsection{A deterministic model of ensemble evolution with first-order trajectory matching}
\label{sec:current_velocity}
We first construct a deterministic dynamical-system model that describes how the stochastic trajectories move locally through state space. 
For $0<t<T$, consider the conditional mean displacement over a short interval centered at $t$,
    \begin{equation}
    \label{eq:symmetric_motion}
    v_\epsilon(t,x) = \mathbb{E}\left[
    \frac{X(t+\epsilon)-X(t-\epsilon)}{2\epsilon}\,\middle|\, X(t)=x\right],
    \end{equation}
    where $0<\epsilon<\min\{t,T-t\}$.
    This averages the motion before and after time $t$ among trajectories passing through the state $x$.
Under suitable regularity assumptions, the limit of this symmetric mean motion is
    \begin{equation}
    \label{eq:localtransport}
    v(t,x) = \lim_{\epsilon\downarrow0}v_\epsilon(t,x) = b(t,x) - \frac{1}{2}\Sigma(t)\nabla_x\log\rho(t,x),
    \end{equation}
    where $\Sigma(t)=A(t)A(t)^\top$ and the expression is understood where $\rho(t,x)>0$.
    The velocity $v$ is the probability-current velocity
    of the stochastic process \cite{nelson1967dynamical,haussmann1986timereversal,
pavliotis2014stochastic}.

The deterministic dynamical-system model given by 
    \begin{equation}
    \label{eq:ODE:CurrentVelocity}
    \frac{\mathrm{d}}{\mathrm{d}r}\widehat{x}(r) =
    v(r,\widehat{x}(r)),
    \qquad
    \widehat{x}(0)\sim\rho(0),
    \end{equation}
    leads to trajectories $(\widehat{x}(r))_{r \in [0, T]}$ that satisfy 
    \begin{equation}\label{eq:MarginalTransportProperty}
    \widehat{x}(r)\sim\rho(r)\,,\qquad r\in[0,T]\,,
    \end{equation}
   see, e.g., \cite{chetrite2009eulerian,boffi2023probabilityflow}. Thus, model \eqref{eq:ODE:CurrentVelocity} generates trajectories that follow both the time marginals and the local mean motion defined above; see \Cref{rmk:MoreThanMarginals}.
     We use FTM~\cite{jha2026ftm} to learn an approximation $v_\theta$ directly from the trajectory data \eqref{eq:TrainingData}, where $\theta$ denotes the model parameters.

\begin{remark}\label{rmk:MoreThanMarginals}
Matching the time marginals specifies where ensemble members are distributed, but not how they evolve over time. For example, clockwise and counterclockwise rotation can preserve the same rotationally symmetric distribution while producing opposite average rotational motion. A dynamical-system model based on the probability-current velocity  \eqref{eq:ODE:CurrentVelocity} additionally matches the average displacement per unit time \eqref{eq:symmetric_motion} of stochastic trajectories passing through each state, measured over a shrinking interval centered at $t$. It therefore captures the direction and rate of this averaged motion, which the time marginals alone do not determine. We refer to \cite{boffi2023probabilityflow,jha2026ftm} for an in-depth discussion. 
\end{remark}

\subsection{Flow maps}
\label{sec:surrogate_flow_maps}
Integrating dynamics \eqref{eq:ODE:CurrentVelocity} defines the flow map
    \begin{equation}
    \label{eq:current_velocity_flow}
    F(x,t,s) = x+ \int_t^s v\bigl(r,F(x,t,r)\bigr)\,\mathrm{d}r, \qquad 0\leq t\leq s\leq T.
    \end{equation}
    Here $F(x,t,r)$ is the state reached at the intermediate time $r$, so the integral accumulates the velocity along the evolving trajectory. Thus, $F(x,t,s)$ gives the state reached at time $s$ when starting from $x$ at time $t$. Note that the marginal-transport property \eqref{eq:MarginalTransportProperty} of the probability-current dynamics induces  that  $F(X(t),t,s)\sim\rho(s)$ if $X(t)\sim\rho(t)$. 
    Thus, applying $F$ to an ensemble distributed according to $\rho(t)$ at time $t$ produces the correct distribution at time $s$.

    Let us now consider the learned $v_\theta$, which analogously induces a flow map 
    \begin{equation}
    \label{eq:learned_velocity_flow}
    F_\theta(x,t,s)
    =
    x+
    \int_t^s
    v_\theta\bigl(r,F_\theta(x,t,r)\bigr)
    \,\mathrm{d}r.
    \end{equation}
    Thus, $F_\theta(x,t,s)$ is the state reached by integrating the learned velocity $v_{\theta}$ from the same initial state $x$ at time $t$.

\subsection{Learning the flow map}
\label{sec:learning_flow_map}
Evaluating $F_\theta(x,t,s)$ by directly using its definition \eqref{eq:learned_velocity_flow} entails numerically integrating $v_\theta$ over the jump horizon $\tau=s-t$.
We now approximate $F_\theta$ with a  parametrized map that can be evaluated without resorting to integrating the velocity $v_{\theta}$; this is a common task in, e.g., machine learning where flow maps of flow-based generative models are learned \cite{boffi2025flowmapmatching,boffi2025selfdistillation,geng2025meanflow}.
Consider 
    \[
    F_\psi: \mathbb{R}^d \times \{(t,s):0\leq t\leq s\leq T\} \rightarrow \mathbb{R}^d,
    \]
    where $\psi$ denotes the parameters to be learned. 
We train $F_\psi$ by minimizing the loss
    \begin{equation}
    \label{eq:Loss:Psi}
    \mathcal{L}(\psi;\theta)
    =
    \mathbb{E}_{\substack{
        t\sim\mathcal{U}([0,T])\\
        s\mid t\sim\mathcal{U}([t,T])
    }}
    \mathbb{E}_{X(t)\sim\rho(t)}
    \left[
    \left\|
    F_\psi(X(t),t,s)
    -
    F_\theta(X(t),t,s)
    \right\|_2^2
    \right],
    \end{equation}
    where $\mathcal{U}$ denotes the uniform
    distribution.
    This objective averages the prediction error
    over starting states and pairs of times.
    A single map is therefore trained over different
    starting times and jump horizons, rather than
    separately for each horizon.

After training, a prediction over $[t,t+\tau]$ with jump horizon $\tau$ is obtained by evaluating $F_\psi(x,t,t+\tau)$. This requires one evaluation of $F_{\psi}$ per starting state and jump, without integrating $v_\theta$ through the intermediate times.

\subsection{Ensemble prediction with the learned flow map}
\label{sec:flow_map_prediction}
Let $0=s_0<s_1<\cdots<s_M=T$ be the requested
    prediction times. We define the jump horizon
    $\tau_k=s_{k+1}-s_k$, which can vary between
    prediction times and can be much larger than
    the time-step size $h$.
For each ensemble member, we sample an initial
    state $z_0\sim\rho(0)$ and sequentially evaluate
    the learned flow map,
    \begin{equation}
    \label{eq:autoregressive_flow_map}
    z_{k+1}
    =
    F_\psi(z_k,s_k,s_k+\tau_k),
    \qquad k=0,\ldots,M-1.
    \end{equation}
    The rollout is autoregressive because each
    predicted state becomes the initial state for
    the next jump horizon.

 Generating one ensemble trajectory requires
    only $M$ flow-map evaluations, compared with
    approximately $T/h$ integration steps for FTM.
    In particular, when $\tau_k\gg h$, FM-FTM
    avoids evaluating the FTM velocity field at
    the intermediate times.

\subsection{Quantities of interest}
\label{sec:qoi}
We consider two types of QoIs for stochastic ensemble prediction. First, QoIs that are ensemble averages of functions of the state and depend only on the time marginals $\rho(t)$.

Second, we consider probability-current QoIs that characterize the local transport of probability mass through state space. Such current-like observables arise naturally in stochastic thermodynamics, where they are used to characterize nonequilibrium transport and dissipation~\cite{seifert2012stochastic,li2019dissipation,frishman2020learning, boffi2024probabilityflows,boffi2025modelfree}.
   Let
    $\phi:[0,T]\times\mathbb{R}^d\to\mathbb{R}^d$
    be a smooth test function. We define the
    time-averaged probability-current QoI as
    \begin{equation}
    \label{eq:current_qoi}
    Q_\phi
    =
    \frac{1}{T}
    \mathbb{E}\left[
    \int_0^T
    \phi(t,X(t))^\top\circ\mathrm{d}X(t)
    \right],
    \end{equation}
    where $\circ$ denotes Stratonovich integration.
    In terms of the probability-current velocity, this quantity can equivalently be written as
    \begin{equation}
    \label{eq:current_qoi_velocity}
    Q_\phi
    =
    \frac{1}{T}
    \int_0^T
    \mathbb{E}\left[
    \phi(t,X(t))^\top v(t,X(t))
    \right]\mathrm{d}t.
    \end{equation}
    Depending on the choice of $\phi$, these QoIs describe quantities such as probability fluxes and circulations.

    The deterministic probability-current dynamics
    \eqref{eq:ODE:CurrentVelocity} preserve both the time marginals and probability-current QoIs of the stochastic process.
    Since the flow map $F$ represents the evolution of these dynamics, it inherits these properties. FM-FTM approximates the flow map induced by the learned FTM velocity and therefore aims to retain the same QoIs, up to the approximation errors of FTM and flow-map learning. In particular, local probability-current
    quantities can be estimated from short-horizon flow-map evaluations without integrating the FTM velocity field.

\begin{figure}
    \centering
    \includegraphics[width=1.0\linewidth]{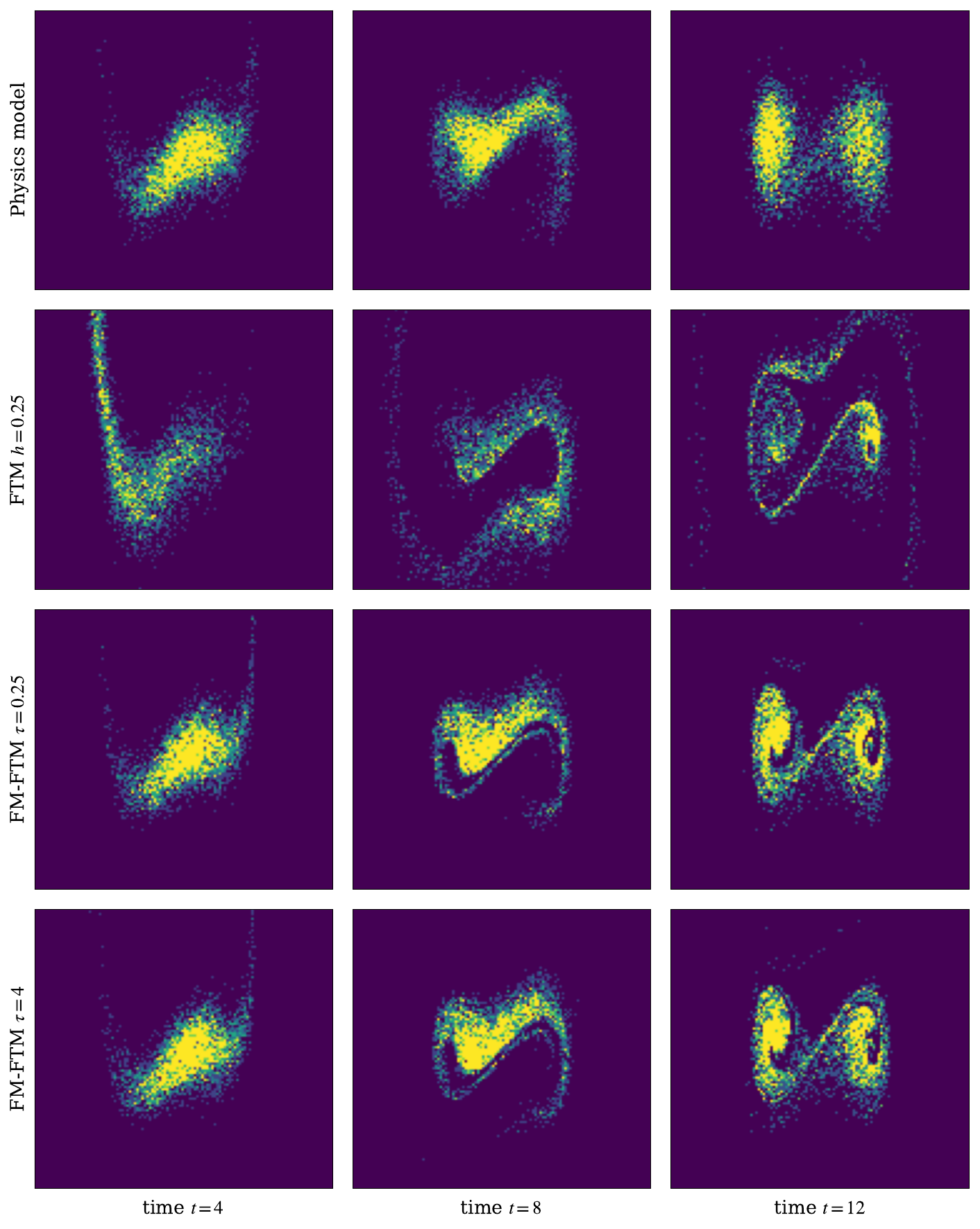}
    \caption{Duffing oscillator: FM-FTM accurately captures the evolving distributions over large time intervals, whereas FTM loses accuracy even with moderately increased time steps from the original $h = 0.01$ to $h = 0.25$.}
    \label{fig:duffing_histogram}
\end{figure}

\begin{figure}
    \centering
    \includegraphics[width=1.0\linewidth]{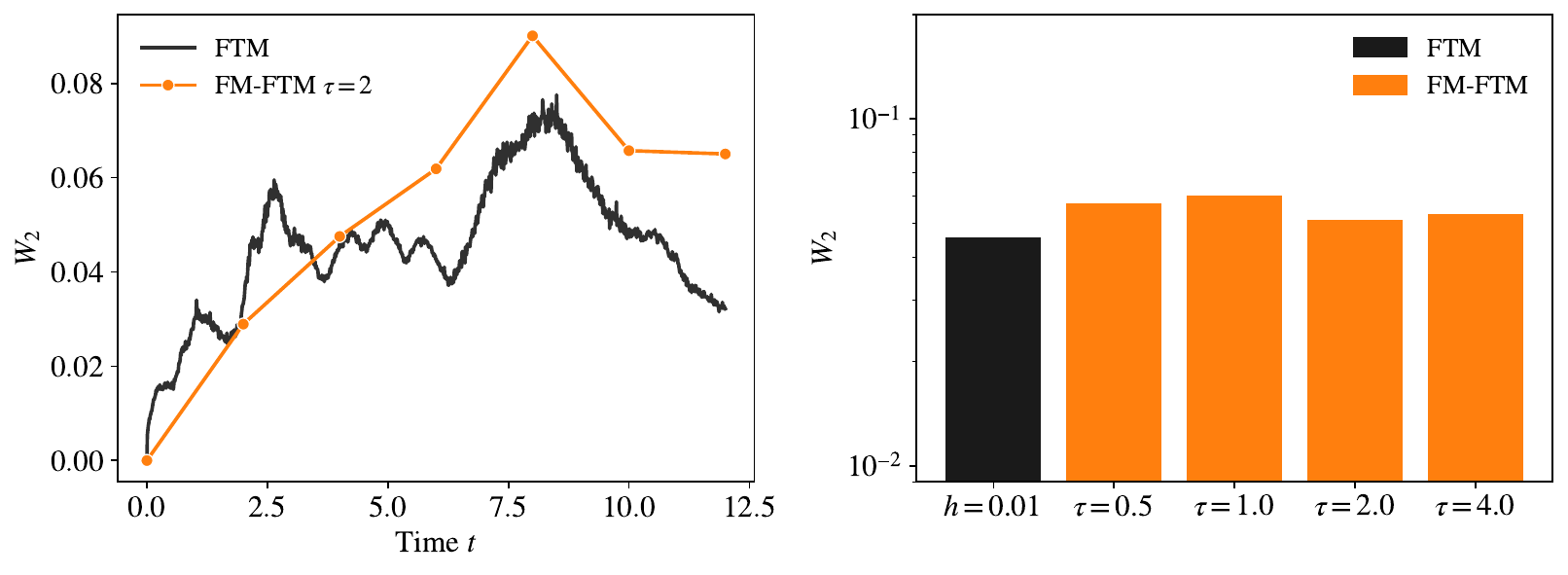}
    \caption{Duffing oscillator: The Wasserstein-2 errors of FM-FTM remain comparable to those of FTM and show little sensitivity to the jump horizon $\tau$, demonstrating that FM-FTM preserves the accuracy of ensemble predictions while reducing the number of time-integration steps by a factor of up to $400\times$ in this example.
}
    \label{fig:duffing_w2}
\end{figure}

\begin{figure}
    \centering
    \includegraphics[width=1.0\linewidth]{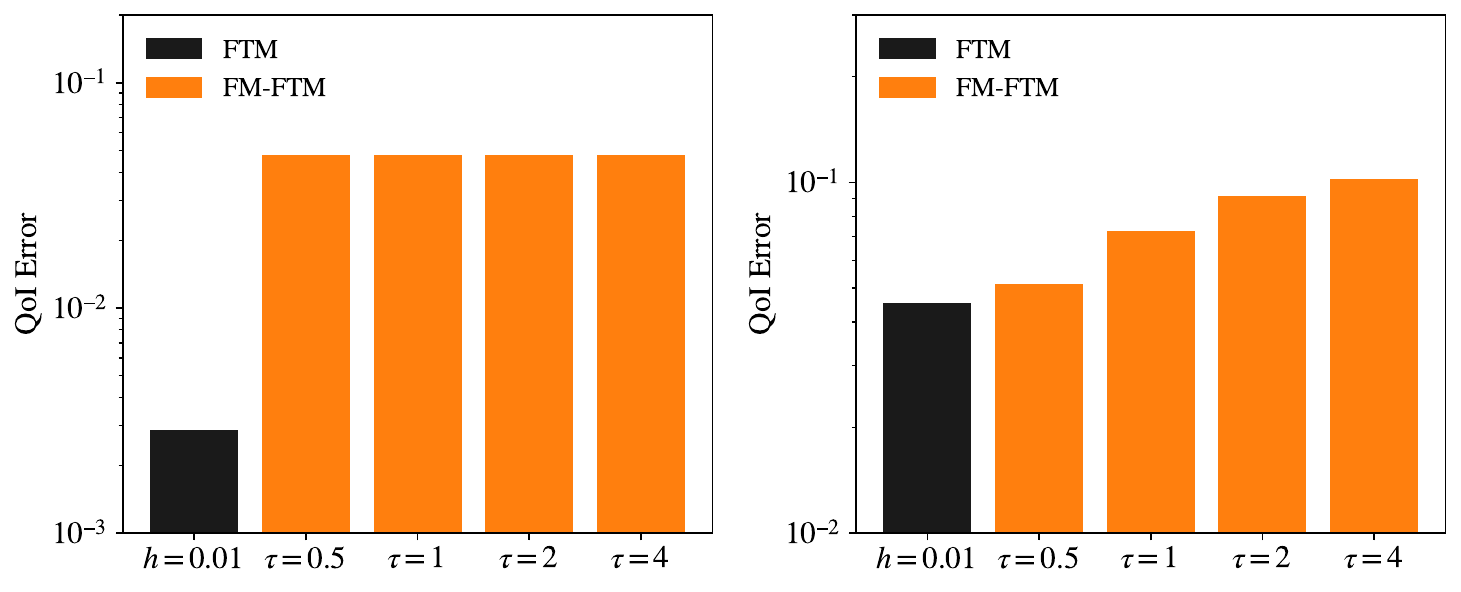}
    \caption{The relative errors in predicting quantities of interest remain small across increasing jump horizons $\tau$ for both the Duffing oscillator (left) and Rayleigh-B\'enard convection (right), demonstrating that FM-FTM preserves trajectory-dependent quantities beyond time-marginal statistics.
}
    \label{fig:qoi_bar_plot}
\end{figure}

\section{Numerical experiments}
We now demonstrate that FM-FTM removes the need to integrate the FTM velocity through each intermediate time step, enabling predictions over large jump horizons $\tau$. 
We evaluate FM-FTM on four different experiments,  Duffing oscillator, Rayleigh-Bénard convection, stochastic Burgers, and stochastic Navier-Stokes.
Our numerical results indicate that FM-FTM preserves time marginals and statistics across tested jump horizons $\tau$ while gaining runtime speedups.

\subsection{Training}
For each system, we first train FTM on the training trajectories to obtain the velocity field $v_\theta$. The architecture used for each system is specified below. We follow  \cite{jha2026ftm} for the FTM training and implementation.
 We then approximate $v_\theta$ with a flow map $F_\psi(x(t), t, s)$ that predicts the state $x(s)$ at times $s$ from the state $x(t)$ at time $t$

 To construct training pairs, we first choose a state $x(t)$ from the training trajectories at time $t$, and then sample $s > t$. Starting from $x_t = x(t)$, we numerically integrate $v_\theta$
    \[
    \frac{\mathrm d }{\mathrm d r} x(r)
    =
    v_\theta(r,x(r)),
    \qquad
    x(t)=x_t,
    \]
    from $t$ to $s$ using the FTM integration step size $h$. We denote the resulting target state by $x_s^\theta$. We then train $F_\psi$ using minibatches of size $B$ by minimizing the empirical mean-squared error
    \[
    \widehat{\mathcal L}(\psi) = \frac{1}{B} \sum_{j=1}^{B} \left\|
    F_\psi(x_{t_j},t_j,s_j)  - x_{s_j}^{\theta} \right\|_2^2.
    \]
 At each training step, $t_j$ is sampled from the time grid of the training trajectories and $s_j$ is sampled from the available grid points strictly after $t_j$.

\subsection{Evaluation Metrics}
For each system, we generate  test trajectories from independently drawn initial conditions using the same solver as for the training data. From the same initial conditions, we roll out FM-FTM at several jump horizons $\tau$ and FTM as a baseline. 
Our evaluation tests three aspects of the learned dynamics. First, the accuracy of predicted time marginals; second, the accuracy of trajectory-dependent probability-current quantities of the form \eqref{eq:current_qoi}; and third, where dimension makes it relevant, the computational cost of ensemble generation. 
To assess time-marginal accuracy, let $X^{(i)}(s_k)$ denote the $i$-th  state and $\widetilde X_k^{(i)}$ the corresponding FM-FTM prediction at time $s_k$. At each rollout time $s_k$, we compute the sliced Wasserstein-2 distance
    
\[
d_k = \operatorname{W}_2
\left(
\{ \widetilde X_k^{(i)} \}_{i=1}^N,
\{ X^{(i)}(s_k) \}_{i=1}^N
\right).
\]
We report the mean sliced Wasserstein-2 distance over the rollout times,
    \[
    \bar d = \frac{1}{M+1} \sum_{k=0}^{M} d_k.
    \]

Agreement in time marginals alone does not imply agreement in how probability mass moves between states (Remark~\ref{rmk:MoreThanMarginals}). We therefore evaluate whether the FM-FTM models learn and can predict trajectory-dependent QoIs.  
For the SDE examples, we focus on the time-local integrand 
    \begin{equation}
    \label{eq:qoi_current}
    q_\phi(t)
    =
    \mathbb{E}\left[
    \phi(X(t))^\top v(t,X(t))
    \right],
    \end{equation}
    where the time-independent test function $\phi$ is specified
    separately for each experiment. Note that $q_{\phi}(t)$ represents the instantaneous contribution to the time-averaged probability-current QoI $Q_{\phi}$. 
    For the rollout times $s_0,\ldots,s_M$,
    with $\tau_k=s_{k+1}-s_k$, we consider the discrete time-weighted average of the time-local probability-current quantities by
    \begin{equation}
    \label{eq:qoi_average}
    \bar{q}_\phi
    =
    \frac{1}{T}
    \sum_{k=0}^{M-1}
    \tau_k q_\phi(s_k).
    \end{equation}
We estimate $\bar{q}_\phi$ from an FM-FTM
    rollout using a midpoint rule.
    Let $\widetilde X_k^{(i)}$ denote the state
    of the $i$-th rollout at time $s_k$.
    From each such state we take one additional
    FM-FTM  evaluation over a small step 
    $\delta_{\mathrm{QoI}}=0.01$, 
    \[
    \widetilde X_k^{+,(i)}
    =
    F_\psi\big(
    \widetilde X_k^{(i)},
    s_k,
    s_k+\delta_{\mathrm{QoI}}
    \big),
    \qquad
    \bar X_k^{(i)}
    =
    \tfrac12
    \big(
    \widetilde X_k^{(i)}
    +
    \widetilde X_k^{+,(i)}
    \big),
    \]
    and compute
    \begin{equation}
    \label{eq:qoi_estimator}
    \widehat q_\phi
    =
    \frac{1}{NT}
    \sum_{i=1}^{N}
    \sum_{k=0}^{M-1}
    \tau_k
    \phi\big(\bar X_k^{(i)}\big)^\top
    \frac{
    \widetilde X_k^{+,(i)}
    -
    \widetilde X_k^{(i)}
    }{
    \delta_{\mathrm{QoI}}
    }.
    \end{equation}
The reference value
    $\widehat q_\phi^{\mathrm{ref}}$
    is computed the same way, with
    $\widetilde X_k^{(i)}$ and
    $\widetilde X_k^{+,(i)}$ replaced by the
    reference states $X^{(i)}(s_k)$ and
    $X^{(i)}(s_k+\delta_{\mathrm{QoI}})$.
    For each jump horizon $\tau$, we report
    the relative error
    $|\widehat q_\phi-
    \widehat q_\phi^{\mathrm{ref}}|
    /|\widehat q_\phi^{\mathrm{ref}}|$.

\subsection{Duffing oscillator with stochastic forcing}
We first consider a stochastically forced duffing oscillator. 
\subsubsection{Duffing oscillator: Setup,  training data, and evaluation metrics}
\label{sec:duffing_set_up}
The dynamics of the duffing oscillator system considered are governed by 
    \[
    \mathrm{d}X_1(t) = X_2(t)\,\mathrm{d}t,
    \qquad
    \mathrm{d}X_2(t)
    =
    \left(-2\xi\omega X_2(t) + \omega^2 X_1(t) - \omega^2\gamma X_1(t)^3\right)\mathrm{d}t
    +
    \sigma\,\mathrm{d}W(t).
    \]
Here $X_1$ denotes the position and $X_2$ the velocity. The parameter $\xi$ controls damping, $\gamma$ the strength of the cubic term, $\omega$ the stiffness of the linear term, and $\sigma$ the strength of the Brownian motion. We use $\xi = 0.2, \gamma = 0.2, \omega=1, \sigma=0.5$.
We integrate this system using the Euler-Maruyama scheme with reference integration step size $h_{\mathrm{ref}}=0.01$ over $t \in [0, 12]$, giving $1200$ time steps per trajectory.
The training data consists of $N=5000$ trajectories generated with initial conditions drawn from $\mathcal{N}([0, -10]^{\top}, I_2)$, where $I_2 \in \mathbb{R}^{2 \times 2}$ is the identity matrix. 

For the FTM velocity model $v_\theta$, we use a three-layer MLP with hidden width $128$ and ReLU activations, with time concatenated to the state input. We follow the training procedure of \cite{jha2026ftm}. For ensemble prediction with FTM, we integrate $v_\theta$ using step size $h=h_{\text{ref}}$.
The flow map $F_\psi$ is parametrized as an MLP with 4 layers, hidden width 256, and ReLU activations. 
The model takes as input the state $x$ at time $t$,
together with the start time $t$ and end time $s$,
and outputs the predicted state $F_\psi(x,t,s)$ at time $s$.
We train with Adam at learning rate $5 \times 10^{-4}$, batch size $2048$, for $500,000$ gradient steps.

We use $5000$ test initial conditions and jump horizons $\tau \in \{0.5, 1.0, 2.0, 4.0\}$. For the probability-current QoI, we use
$\phi(x)=\varphi(x_1)e_1$ with
$\varphi(z)=\frac{1}{\sqrt{2\pi}\epsilon}
e^{-z^2/(2\epsilon^2)}$ and $\epsilon=1$.
This QoI measures the time-averaged, signed probability
flux in the $x_1$-direction near $x_1=0$,
with $\epsilon$ controlling the width of the region
over which the flux is measured.

\subsubsection{Duffing oscillator: Results}
\label{sec:duffing_results}
Figure~\ref{fig:duffing_histogram} compares the reference distribution, FTM evaluated with a coarse integration step $0.25$, and FM-FTM evaluated with jump horizons $\tau=0.25$ and $\tau=4.0$ at $t = 4, 8, 12$. At each snapshot, FTM with a coarse time step fails to predict an accurate marginal distribution, whereas FM-FTM with both jump horizons captures the main distributional structure. Thus, FM-FTM can advance over up to $400\times$ larger jump horizons than the time-step size $h = 0.01$ while remaining accurate, whereas FTM loses accuracy alrady when integrated with a coarser time step of $0.25$.

Figure~\ref{fig:duffing_w2} quantifies the time-marginal accuracy. FM-FTM achieves errors comparable to FTM integrated with step size $h=0.01$, with little sensitivity to the jump horizon across $\tau\in\{0.5,1.0,2.0,4.0\}$. Thus, FM-FTM maintains accurate time marginals with jump horizons up to $400$ times larger than the FTM integration step.

Figure~\ref{fig:qoi_bar_plot} (left) shows that FM-FTM accurately predicts the probability-current QoI across all tested jump horizons, with little sensitivity to $\tau$. In particular, FM-FTM captures the  information about the local transport of probability mass as FTM and not only the time marginals,
even when advancing over large time intervals.

\begin{figure}
    \centering
    \includegraphics[width=1.0\linewidth]{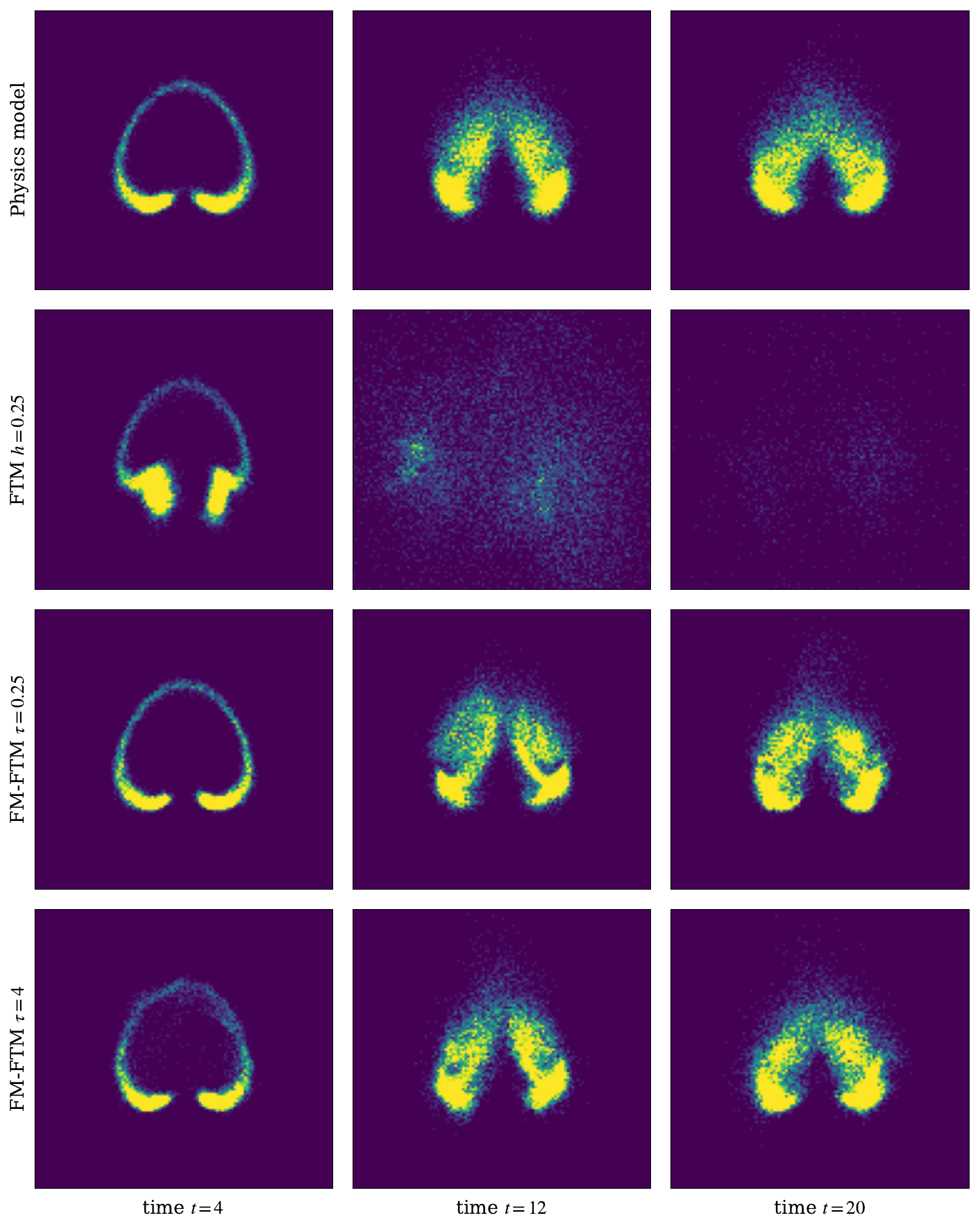}
    \caption{Rayleigh-B\'enard: FTM fails to capture the multimodal distribution even already with moderately increased time-step sizes, whereas FM-FTM accurately captures the evolution of both modes at a jump horizon of $\tau=4$, which is $400$ times larger than the time-step size used by FTM.
}
    \label{fig:lorenz_histogram}
\end{figure}

\subsection{Rayleigh-B\'{e}nard convection with stochastic forcing}
We now consider the stochastically forced nine-dimensional model of Rayleigh-B\'{e}nard convection introduced in \cite{reiterer1998nine}. The underlying deterministic dynamics exhibit chaotic behavior, providing a more challenging setting for ensemble prediction.

\subsubsection{Rayleigh-B\'{e}nard: Setup, training data, and evaluation metrics}
\label{sec:rb_set_up}
We consider the nine-dimensional system governed by
    \[
    \mathrm{d}X(t) = f_\mu(X(t))\,\mathrm{d}t + \sigma\,\mathrm{d}W(t),
    \]
    where $f_\mu: \mathbb{R}^9 \to \mathbb{R}^9$ is the deterministic Rayleigh-B\'enard convection vector field given in \cite{reiterer1998nine} and $\mu$ is the Rayleigh-type control parameter. We use $\mu = 13.65$, and $\sigma = 0.05$. 
    
For the training data, we generate $N = 20000$ trajectories with initial conditions drawn from $\mathcal{N}(0, (0.02)^2 I_9)$, integrated with the Euler-Maruyama scheme using reference integration step size $h_{\mathrm{ref}}=0.01$ over time interval $[0, 20]$.
    
For $v_\theta$, we use a seven-layer MLP with hidden width $128$ and ReLU activations, with time concatenated to the state input. Training follows \cite{jha2026ftm}. For ensemble prediction with FTM, we integrate $v_\theta$ using step size $h=h_{\text{ref}}$.
    
The flow map $F_\psi$ is parametrized as an MLP with 6 layers and hidden dimension 512, with ReLU activations. The input-output structure is the same as in Section \ref{sec:duffing_set_up} for the Duffing oscillator. We train the model with Adam at a learning rate of $5 \times 10^{-4}$, batch size $256$, for $500,000$ gradient steps.

We use $20000$ test initial conditions and jump horizons $\tau \in \{0.5, 1.0, 2.0, 4.0\}$.  For the probability-current QoI, we use
$\phi(x)=-x_8e_6+x_6e_8$.
This QoI measures the time-averaged, signed rotational
transport of probability mass around the origin in the
$(x_6,x_8)$ plane, distinguishing clockwise from
counterclockwise motion.

\subsubsection{Rayleigh-B\'{e}nard: Results}
Figure~\ref{fig:lorenz_histogram} shows that FM-FTM
    accurately captures the evolution of the multimodal
    distribution at $t=4,12,20$, even with a jump horizon
    of $\tau=4$. In contrast, FTM with a coarse integration
    step $0.25$ fails to capture the
    distributional structure. Thus, FM-FTM can advance over
    larger time intervals without losing the multimodal
    features of the reference distribution.

\begin{figure}
    \centering
    \includegraphics[width=1.0\linewidth]{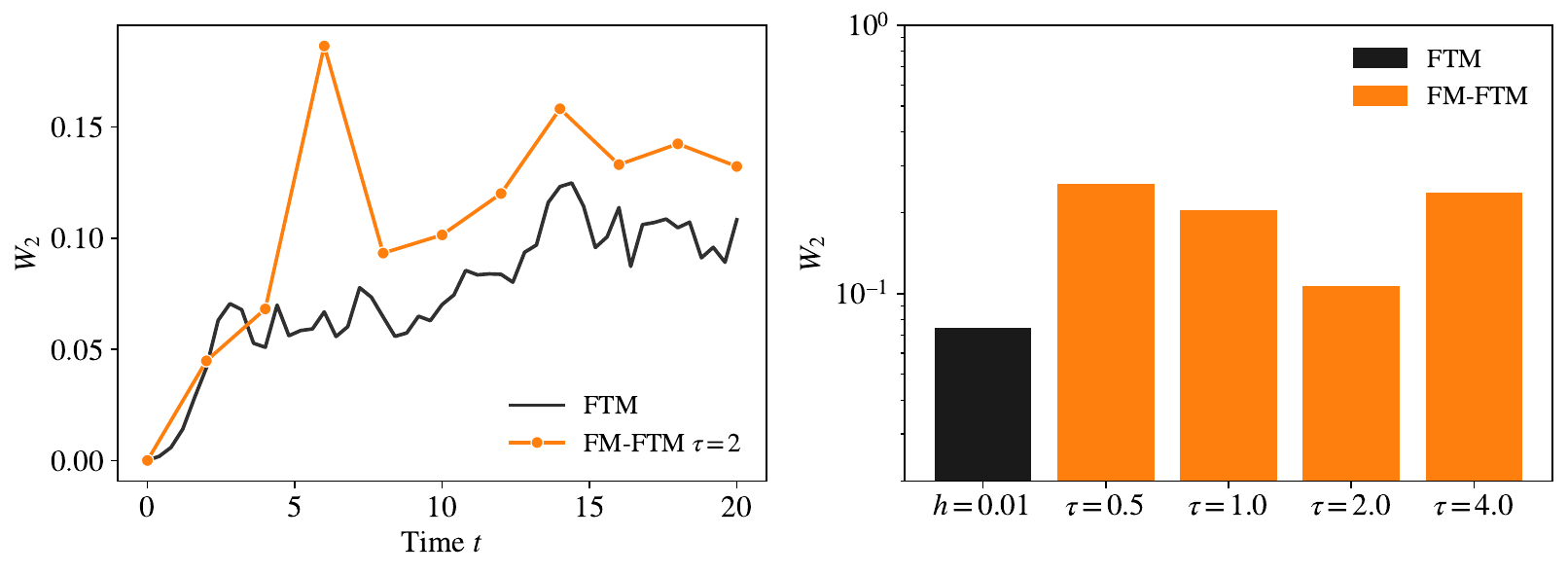}
    \caption{Rayleigh-B\'{e}nard: The Wasserstein-2 errors remain moderate across jump horizons up to $\tau=4$, corresponding to $400$ times the time-step size of FTM.}
    \label{fig:lorenz_w2}
\end{figure}

Figure~\ref{fig:lorenz_w2} quantifies the
    time-marginal accuracy. FM-FTM maintains moderate
    sliced Wasserstein-2 errors across the tested jump
    horizons, with no systematic deterioration as $\tau$
    increases. Although its errors are higher than those
    of FTM integrated with $h=0.01$, FM-FTM retains
    useful marginal accuracy for jump horizons up to
    $400$ times larger than the FTM integration step.

Figure~\ref{fig:qoi_bar_plot} (right) shows that
    FM-FTM accurately predicts the probability-current QoI
    across all tested jump horizons. Thus, FM-FTM retains
    information about the rotational transport of probability
    mass beyond the time marginals, even when advancing
    over large time intervals.

\subsection{Stochastically forced Burgers equation}
We now consider the one-dimensional Burgers equation with
stochastic forcing to evaluate FM-FTM for spatially distributed
dynamics. In this setting, we assess whether FM-FTM accurately
predicts energy and enstrophy of ensemble members over large jump
horizons.

\subsubsection{Burgers: Setup and training data}
We consider the one-dimensional Burgers equation with periodic boundary conditions on spatial domain $[0, 1)$,
    \[
    \partial_tu(t,x)
    =
    \nu\,\partial_{xx}u(t,x) - u(t,x)\,\partial_x u(t,x)
    \]
    with viscosity $\nu=0.007$. We discretize in space on a uniform 64-point grid, yielding an ODE with a state in $\mathbb{R}^{64}$, to which we then add stochastic forcing. Specifically, for each grid point $\zeta_j$, we set 
    \[
    \mathrm{d}W_j(t)
=
\sum_{\iota=1}^{10}
\frac{1}{\iota}
\left(
a_\iota(\zeta_j)\,\mathrm{d}B_\iota^{(1)}(t)
+
b_\iota(\zeta_j)\,\mathrm{d}B_\iota^{(2)}(t)
\right),
    \]
    where each $B_\iota^{(1)}, B_\iota^{(2)}$ represents independent Brownian motions and $a_\iota, b_\iota$ are sine and cosine basis functions. The noise amplitude is $\sigma = 0.04$. We approximate the spatial derivatives with second-order centered finite differences. We then integrate the resulting stochastic system in time with Euler-Maruyama using reference integration step size $h_{\mathrm{ref}}=0.0005$ over $[0,4]$.

    We draw $4096$ initial conditions for our training trajectories  from Gaussian bumps with the same noise structure as the stochastic forcing,
\[
u(0,x)
=
\exp\left(-20(x-0.5)^2\right)
+
0.015\eta(x),
\]
where $\eta$ is a zero-mean Gaussian random field constructed using the same spatial basis as the stochastic forcing. For training, we downsample the $8000$ time steps to $800$ snapshots, meaning our training data has a snapshot spacing of $\Delta t_{\mathrm{obs}}=0.005$. Thus our training data consists of $4096$ trajectories with $800$ timesteps each. 

The FTM velocity model is a one-dimensional U-Net with channel widths $[32,64,128]$, two residual blocks at each resolution, and a middle residual block, using GeLU activations. Time is encoded by an MLP and injected into the residual blocks through learned projections. The rest of the training procedure is as described in \cite{jha2026ftm}. For ensemble prediction with FTM, we integrate $v_\theta$ using step size $h=h_{\text{ref}}$.
    
The flow map $F_\psi$ is a one-dimensional U-Net with channel widths $[32, 64, 128]$, two residual blocks per scale, one middle residual block, and GeLU activations. The start and end times $t,s$ are concatenated and embedded via a 2-layer MLP of width 128, then injected into each residual block via a learned linear projection. The projected vector is added to the feature map before both convolutions in the block, broadcast across the spatial dimension. 

In training the network, the optimizer used is Adam, gradient-clipped to global norm $1.0$, with a warm-up cosine schedule and peak learning rate $5 \times 10^{-4}$. The network is trained for $1,000,000$ training steps with batch size $64$.

\subsubsection{Burgers: Evaluation metrics.}
We use $4096$ test initial conditions and jump horizons $\tau \in  \{0.25, 0.5, 1.0\}$. To assess marginal accuracy, we track energy and enstrophy, 
 \[
    E(t) = \tfrac12\int_0^1 u(t,x)^2\,\mathrm dx,
    \qquad
    Z(t) = \tfrac12\int_0^1 u_x(t,x)^2\,\mathrm dx,
\]
which we estimate on the periodic grid using a second-order finite difference $u_{x,j}$. Thus, our sample estimators are
    \[
    \widehat E(t) = \tfrac12\sum_j u_j(t)^2\,\mathrm dx,
    \qquad
    \widehat Z(t) = \tfrac12\sum_j u_{x,j}(t)^2\,\mathrm dx.
    \]
At all time steps, we average $\widehat E(t)$ and $\widehat Z(t)$ over all $4096$ trajectories for both the flow map and reference samples, and compute the relative error between the two. We plot these relative errors over time.

\subsubsection{Burgers: Results}
\begin{figure}
    \centering
    \includegraphics[width=1.0\linewidth]{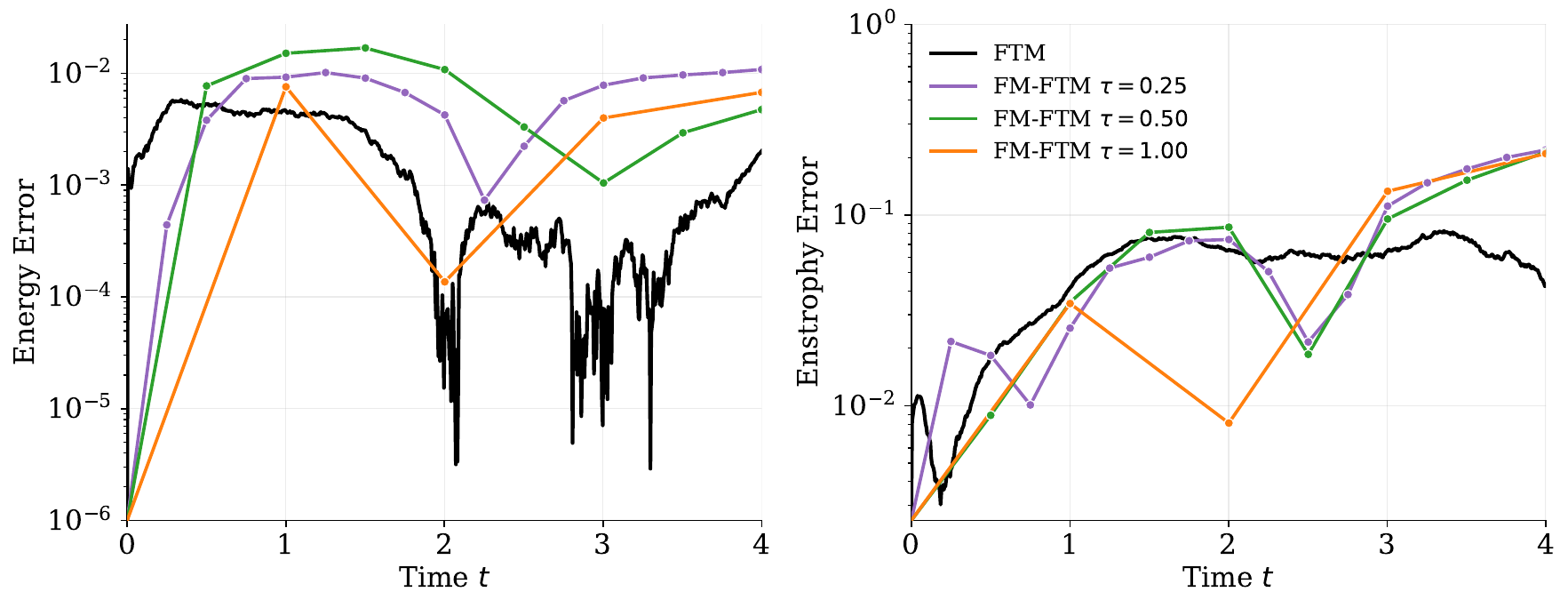}
    \caption{Burgers: The relative errors in energy (left) and enstrophy (right) exhibit similar behavior for FTM and FM-FTM across different jump horizons $\tau$, demonstrating that FM-FTM retains the accuracy of these physical ensemble statistics while reducing the number of time-integration steps.
}
    \label{fig:burgers_energy_enstrophy}
\end{figure}

Figure~\ref{fig:burgers_energy_enstrophy} shows that
    FM-FTM accurately predicts ensemble-mean energy across
    all tested jump horizons $\tau\in\{0.25,0.5,1.0\}$.
    The relative errors remain below $2\%$ throughout the
    prediction interval. Thus, FM-FTM retains accurate energy
    predictions even when replacing multiple integration
    steps with a single flow-map evaluation. FM-FTM also captures the evolution of enstrophy across the tested jump horizons. Enstrophy errors are generally larger than energy errors and exceed those of FTM at later times.
    These results demonstrate that FM-FTM can predict physical ensemble statistics over large jump horizons.

\subsection{Forced turbulence}
We consider the two-dimensional incompressible Navier-Stokes equations with stochastic forcing.

\subsubsection{Forced turbulence: Setup and training data}
Consider the incompressible Navier-Stokes equation in vorticity form on the periodic domain
    $\Omega=[0,2\pi)^2$. The viscosity is $\nu=10^{-3}$ and the linear
    drag coefficient is $\alpha=0.1$. The stochastic forcing acts through
    low-frequency Fourier modes
    \[
    \mathcal{K}_{\mathrm{sto}}
    =
    \{\kappa\in\mathbb{Z}^2:0<|\kappa|\leq4\}.
    \]
    At spatial grid point $\zeta_j$, the stochastic
    increment is
    \[
    \mathrm{d}W_j(t)
    =
    \sigma
    \sum_{\kappa\in\mathcal{K}_{\mathrm{sto}}}
    w_\kappa
    \left[
    \cos(\kappa\cdot \zeta_j)\,
    \mathrm{d}B_\kappa^{(1)}(t)
    +
    \sin(\kappa\cdot \zeta_j)\,
    \mathrm{d}B_\kappa^{(2)}(t)
    \right],
    \]
    where the Brownian motions are independent,
    $w_\kappa\propto|\kappa|^{-1/2}$, and
    $\sigma=0.3$.  Initial conditions are sampled from the
     initial-condition ensemble given in 
    \cite{kochkov2021machinelearning}
    and rescaled so that the maximum vorticity is seven.
    We simulate the reference dynamics on a
    $256\times256$ grid using a pseudo-spectral
    discretization with $2/3$-rule dealiasing.
    The linear terms are integrated with
    Crank-Nicolson and nonlinear advection
    with a fourth-order Runge-Kutta scheme.

  We generate $N=2048$ training trajectories over time interval $[0,25]$ using reference integration step size $h_{\mathrm{ref}}=0.001$.
    Each trajectory is temporally subsampled
    to $1000$ snapshots and spectrally
    restricted to a $64\times64$ spatial grid
    for training.

    The FTM velocity model $v_\theta$ is a
    two-dimensional U-Net with channel widths
    $[128,256,512,1024]$, two residual blocks
    per resolution, and a middle residual block,
    using GeLU activations.
    The model is conditioned on three consecutive
    state frames, with time embeddings injected
    through learned projections.
    We follow~\cite{jha2026ftm} for the remaining
    training details and integrate $v_\theta$
    with step size $h=0.025$ for ensemble prediction.

   The flow map $F_\psi$ uses the same U-Net
    architecture, with the state as convolutional
    input and start and end times $(t,s)$ embedded
    through an MLP and injected into each residual
    block.
    We train for $1{,}000{,}000$ gradient steps
    with batch size $128$ using Adam, gradient
    clipping at norm $1.0$, and a warm-up cosine
    schedule with peak learning rate $10^{-4}$.

\begin{figure}
    \centering
    \includegraphics[width=1.0\linewidth]{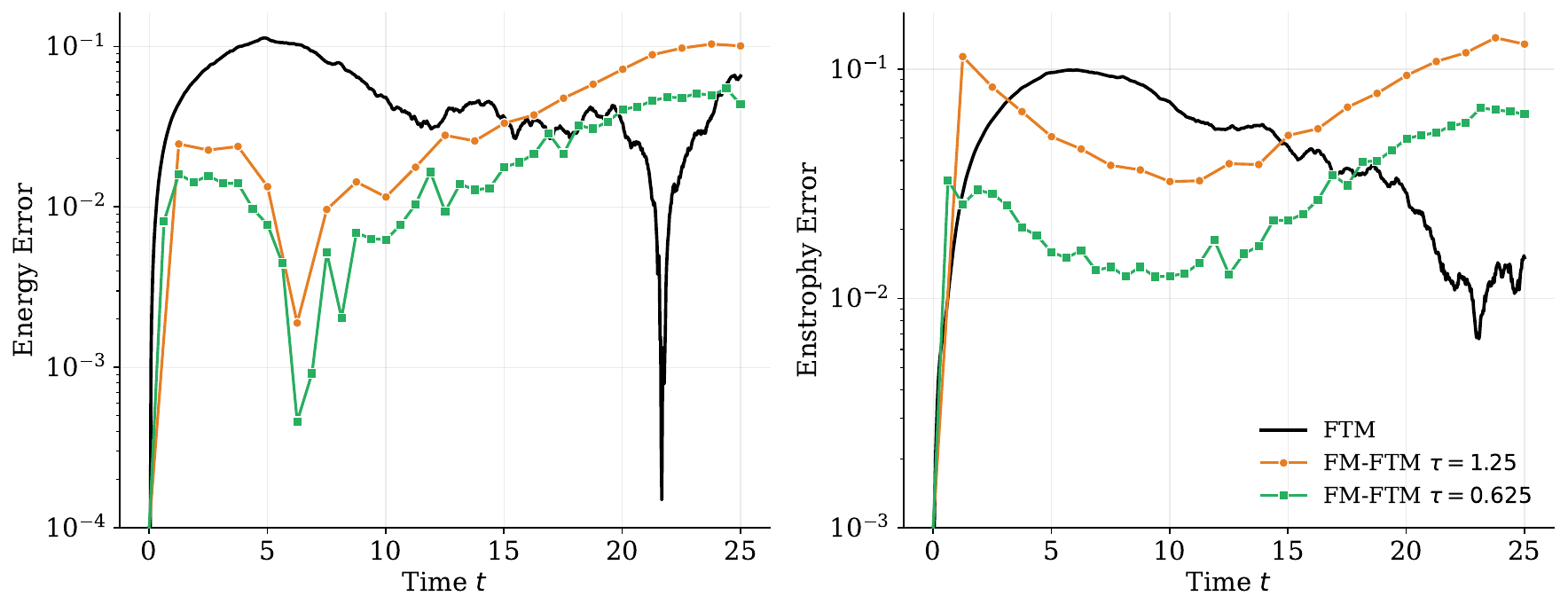}
    \caption{Stochastically forced turbulence: Over the time interval, FM-FTM requires only $20$ ($\tau=1.25$) flow-map evaluations, compared with $1000$ sequential integration steps for FTM ($h=0.025$), while achieving comparable relative errors in ensemble-mean energy (left) and enstrophy (right).}
    \label{fig:ns_energy_enstrophy}
\end{figure}

\begin{figure}
    \centering
    \includegraphics[width=1.0\linewidth]{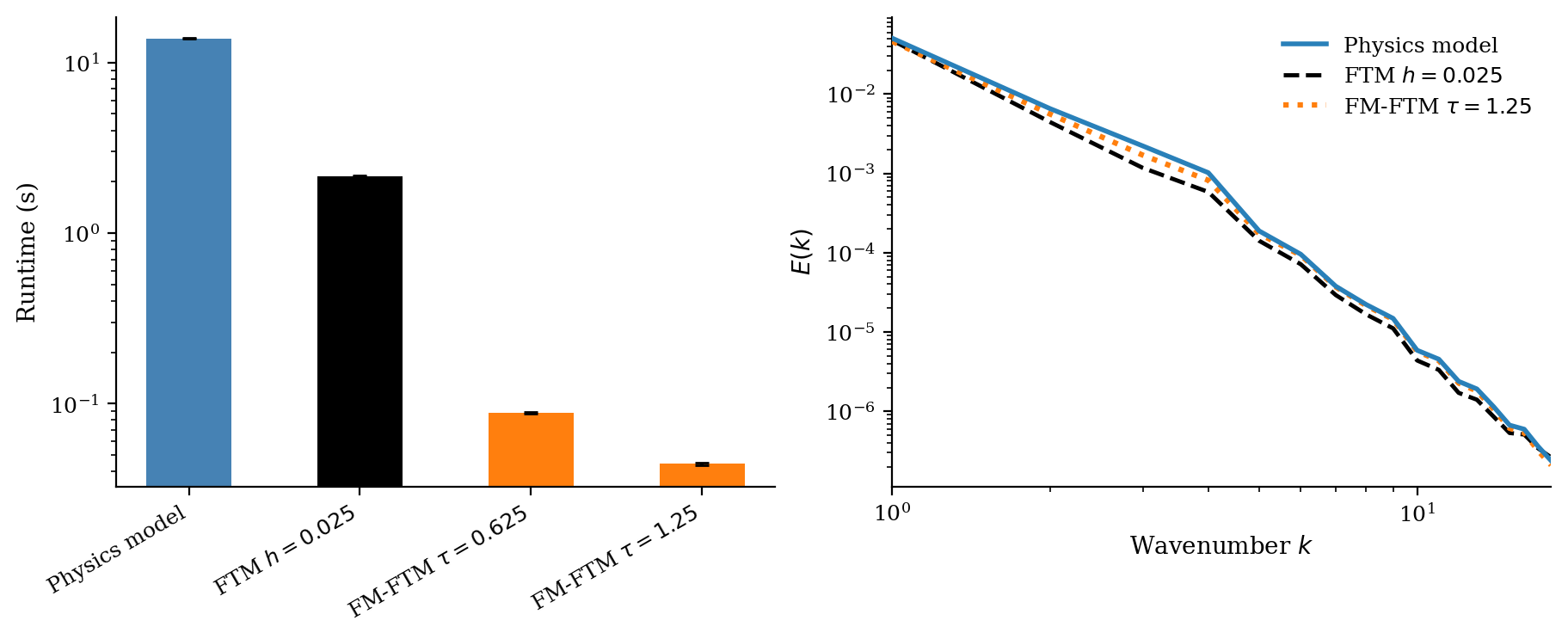}
    \caption{Stochastically forced turbulence: With a jump horizon of $\tau=1.25$, FM-FTM generates trajectories approximately $50\times$ faster than FTM and about $500\times$ faster than the physics model (left), while predicting solution fields that accurately match the decay of the kinetic-energy spectrum across wavenumbers (right).}
    \label{fig:ns_energy_spectrum}
\end{figure}

\subsubsection{Turbulence: Evaluation metrics}
We use $1024$ test initial conditions
    and FM-FTM jump horizons
    $\tau\in\{0.625,1.25\}$.
    We evaluate the relative errors in ensemble-mean
    kinetic energy and enstrophy, computed from
    the velocity and vorticity fields, respectively.
    We also compare the kinetic-energy spectra
    at selected times to assess the distribution
    of energy across spatial scales.
To assess computational efficiency,
    we measure the wall-clock time required
    to generate trajectories with the reference
    physics model, FTM, and FM-FTM at the
    tested jump horizons.

\subsubsection{Forced turbulence: Results}
Integrating the learned FTM velocity field becomes unstable when the time-step size is increased from $h = 0.025$ to $h = 0.25$, as shown in Figure~\ref{fig:navier_stokes_grid}.  In contrast, FM-FTM produces accurate predictions with jump horizon $\tau=0.25$. Thus, FM-FTM enables rollouts over large jump horizons while maintaining accurate predictions and preserving coherent vorticity structures in this example. Accurate ensemble statistics are retained at even larger jump horizons, as shown in Figure~\ref{fig:ns_energy_enstrophy}, where FM-FTM accurately predicts  kinetic energy and enstrophy across the tested jump horizons $\tau\in\{0.625,1.25\}$, with relative errors of similar magnitude to those of FTM over much of the time interval. With $\tau=1.25$, FM-FTM requires only $20$ flow-map evaluations over time interval $[0,25]$, compared with $1000$ sequential integration steps for FTM using $h=0.025$. Thus, FM-FTM retains accurate predictions of physical ensemble statistics while reducing the number of sequential evaluations by a factor of $50$.

Figure~\ref{fig:ns_energy_spectrum} (right) shows that FM-FTM with $\tau=1.25$ also predicts well the kinetic-energy spectrum at $t=25$, closely matching the reference across the plotted wavenumbers. This demonstrates that FM-FTM captures not only integrated ensemble statistics but also the distribution of kinetic energy across spatial scales.

Figure~\ref{fig:ns_energy_spectrum} (left) demonstrates the computational savings. FM-FTM reduces trajectory-generation time by approximately $26\times$ and $49\times$ relative to FTM for $\tau=0.625$ and $\tau=1.25$, respectively. Compared with the physics model, FM-FTM with $\tau=1.25$ accelerates predictions by more than two orders of magnitude. These results provide further evidence that FM-FTM reduces the computational cost of ensemble prediction while retaining accurate turbulence statistics.

\section{Conclusion}
Our results demonstrate that stochastic ensemble predictions can be accelerated by reducing the dynamics in time, rather than only reducing the state dimension or making individual time steps cheaper. By approximating the evolution of a learned probability-current velocity with a flow map, FM-FTM replaces sequential time integration with evaluations of a learned flow map. Our numerical experiments show that FM-FTM accurately captures time marginals, probability-current
quantities, and physical ensemble statistics over time horizons that are orders of magnitude larger than the time-integration step size of the underlying dynamical model. Thus, overall, these results show that the time resolution needed to accurately integrate a dynamical model need not determine the computational cost of predictions when predictions are required only at selected times. This puts model reduction in time, via, e.g., flow-map approximations as in this work, as a complementary approach to  model reduction in state space.

\bibliography{main}

\end{document}